\documentclass[runningheads]{llncs}
\usepackage{graphicx}
\usepackage{longtable,rotating}
\usepackage{threeparttable}

\newcommand\blfootnote[1]{%
  \begingroup
  \renewcommand\thefootnote{}\footnote{#1}%
  \addtocounter{footnote}{-1}%
  \endgroup
}

\usepackage[hyphens]{url}

\begin{document}

\title{mRAT-SQL+GAP: \\ A Portuguese Text-to-SQL Transformer\thanks{Supported by IBM and FAPESP (São Paulo Research Foundation).}}
%
%
\author{Marcelo Archanjo José\inst{1}\orcidID{0000-0001-7153-0402} \and\\
Fabio Gagliardi Cozman\inst{2}\orcidID{0000-0003-4077-4935}}
%
%

\institute{Center for Artificial Intelligence (C4AI) and \\
Instituto de Estudos Avançados, Universidade de São Paulo, Brazil \\ \email{marcelo.archanjo@usp.br} \and
Escola Politécnica, Universidade de São Paulo and \\ 
Center for Artificial Intelligence (C4AI), Brazil \\ \email{fgcozman@usp.br}}

\maketitle              
\begin{abstract}
The translation of natural language questions to SQL queries has attracted 
growing attention, in particular in connection with transformers and similar 
language models.
A large number of techniques are geared towards the English language; in this 
work, we thus investigated translation to SQL when
input questions are given in the Portuguese language.
To do so, we properly adapted state-of-the-art tools and resources. We changed
the RAT-SQL+GAP system by relying on a multilingual BART model (we report
tests with other language models), and we produced a translated 
version of the Spider dataset.
Our experiments expose interesting phenomena that arise when
non-English languages are targeted; in particular, it is better to train
with original and translated training datasets together, {\em even} if a
single target language is desired. 
This multilingual BART model fine-tuned with a double-size training dataset (English and Portuguese) achieved 83\% of the baseline, making inferences for the Portuguese test dataset. This investigation can help other researchers to produce results in Machine Learning in a language different from English. Our multilingual ready version of RAT-SQL+GAP and the data are available, open-sourced as mRAT-SQL+GAP at:  \url{https://github.com/C4AI/gap-text2sql}.\blfootnote{BRACIS 2021, LNAI 13074, pp. 511-525, 2021. The final authenticated version is available online at \url{ https://doi.org/10.1007/978-3-030-91699-2_35}}

\keywords{NL2SQL \and Deep Learning \and RAT-SQL+GAP \and Spider dataset \and BART \and BERTimbau.}
\end{abstract}

\section{Introduction}

\begin{figure}[t]
\includegraphics[width=\textwidth]{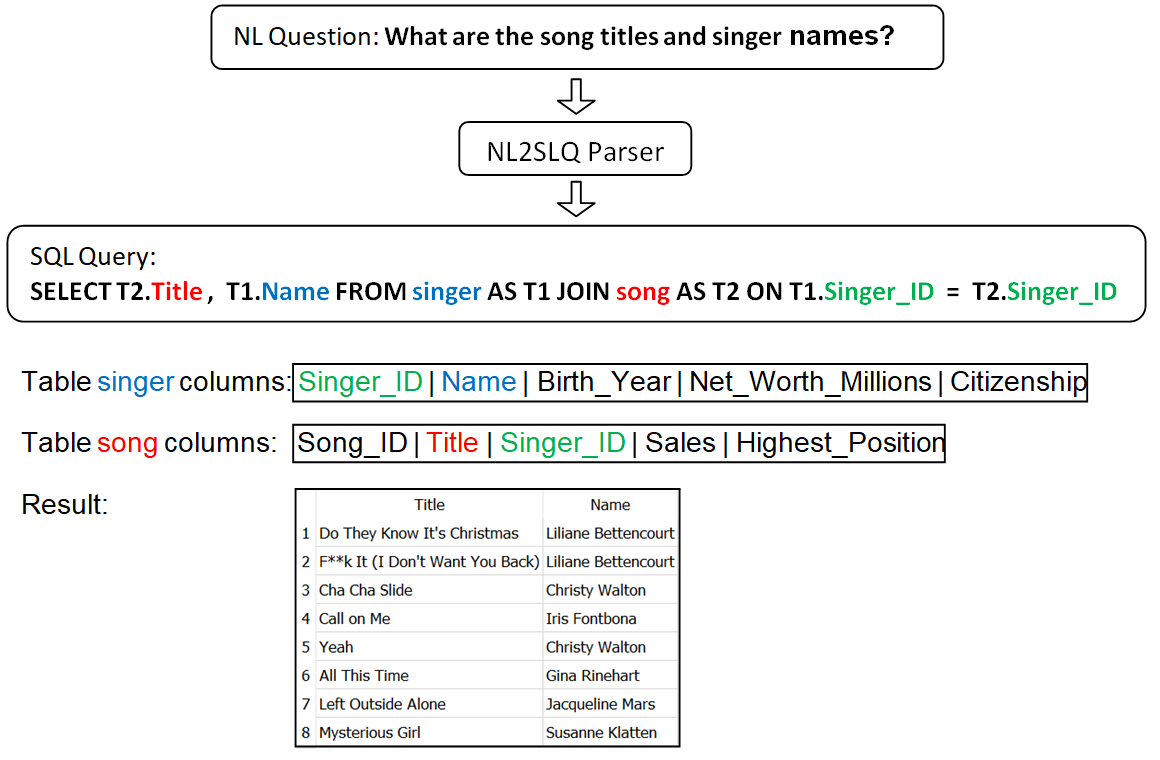}
\caption{From a natural language question to a SQL query and to the database query result.
Database names appear in blue and red, while the primary key appears in green.} 
\label{fig1}
\end{figure}

A huge number of data is now organized in relational databases and 
typically accessed through SQL (Structured Query Language) queries.
The interest in automatically translating questions expressed in 
natural language to
SQL (often referred to as NL2SQL) has been intense, as one can
observe through a number
of excellent surveys in the literature~\cite{Kim2020,Affolter2019,Ozcan2020}.
Fig.~\ref{fig1} depicts the whole flow from a natural language question
to a SQL query result; the SQL query refers to database tables
and their columns, using primary and secondary keys as appropriate.

Existing approaches for NL2SQL can be divided into entity-based and machine 
learning ones, the latter dominated by techniques based on deep learning~\cite{Ozcan2020}.

Entity-based approaches focus on the interpretation of input text based on rules so as to translate it to a SQL query. The translation often goes first to an intermediary state and later to a final SQL query. Relevant systems are Bela~\cite{Walter2012}, SODA~\cite{Blunschi2012}, NaLIR~\cite{Li2014,Li2014a,Li2016}, TR Discover~\cite{Song2015}, Athena~\cite{Saha2016,Lei2018} , Athena++~\cite{Sen2020} and Duoquest~\cite{Baik2020,Baik2020a} . 

Machine learning approaches are based on supervised learning, in which training data contains 
natural language questions and  paired SQL queries~\cite{Ozcan2020}. Several 
architectures can be trained or fine-tuned so as to run the translation.
Relevant systems are EchoQuery~\cite{Lyons2016}, Seq2SQL~\cite{Zhong2017}, 
SQLNet~\cite{Xu2017}, DialSQL~\cite{Gur2018}, TypeSQL~\cite{Yu2018}, 
SyntaxSQLNet~\cite{Yu2018a}, AugmentedBI~\cite{Francia2019}, IRNet~\cite{Guo2019}, 
RAT-SQL~\cite{Wang2019}, RAT-SQL+GAP~\cite{Shi2020}, GraPPa~\cite{Yu2020}, 
BRIDGE~\cite{Lin2020}, DBPal~\cite{Utama2018,Basik2018,Weir2020},
HydraNet~\cite{Lyu2020}, DT-Fixup~\cite{Xu2021} and LGESQL~\cite{Cao2021}. 

There are also hybrid approaches that combine entity-based and machine learning~\cite{Ozcan2020}; relevant examples are Aqqu~\cite{Bast2015}, MEANS~\cite{BenAbacha2015} and QUEST~\cite{Bergamaschi2016}.

The previous paragraphs contain long lists of references that should suffice
to demonstrate that translation from natural language to SQL is a well explored
research topic. The field is relatively mature and 
benchmarks for NL2SQL are now widely used, containing
training and testing data and ways to evaluate new proposals. 
Table~\ref{tab1} shows a few  important datasets in the literature,
reporting their number of questions, number of SQL queries, 
number of databases, number of domain and tables per database \cite{Yu2018b}.
Particularly, the Spider dataset is a popular resource that contains
200 databases with multiple tables under 138 domains.\footnote{Spider 
dataset: \url{https://yale-lily.github.io/spider}.} 
The complexity of these tables allows testing complex nested SQL queries.
A solid test suite evaluation package for testing against Spider \cite{Zhong2020}
is available;\footnote{Spider test suite evaluation github:\url{https://github.com/taoyds/test-suite-sql-eval}.}
in addition, there is a very active leaderboard rank for tests that use
Spider.\footnote{Spider leaderboard rank: \url{https://yale-lily.github.io/spider}.} 

\begin{table}[t]
\centering
\caption{NL2SQL Benchmarks.}\label{tab1}
\begin{tabular}{|l|l|l|l|l|l|l|}
\hline
Dataset &  Questions & SQL Queries & Databases & Domains & Tables/DB & References\\
\hline
ATIS & 5,280 & 947 & 1 & 1 & 32 & \cite{Iyer2017,Zettlemoyer2007,Price1990,Dahl1994,Hemphill1990}\\
GeoQuery & 877 & 247 & 1 & 1 & 6 & \cite{Zettlemoyer2005,Zelle1996,Iyer2017,Giordani2012,Popescu2003}\\
WikiSQL & 80,654 & 77,840 & 26,521 & * & 1 & \cite{Zhong2017}\\
Spider & 10,181 & 5,693 & 200 & 138 & 5.1 & \cite{Yu2018b}\\ 
\hline
\end{tabular}
\begin{tablenotes}
\item * The WikiSQL has multiple domains, but the organization of one table per database does not allow exploring the complexity of the different domains.
\end{tablenotes}
\end{table}

Currently, the best result in the Spider leaderboard for \emph{Exact set match without values}, whereby a paper and code are available, 
is the entry by RAT-SQL+GAP \cite{Shi2020}.\footnote{RAT-SQL+GAP gitHub: \url{https://github.com/awslabs/gap-text2sql}.} This system appears in 
the 6th rank position with Dev 0.718 and Test 0.697.\footnote{Dev results are obtained locally by the developer; to get official score and Test results, it is necessary to submit the model following guidelines in ``Yale Semantic Parsing and Text-to-SQL Challenge (Spider) 1.0 Submission Guideline'' at \url{ https://worksheets.codalab.org/worksheets/0x82150f426cb94c17b861ef4162817399/}.}
Note that the Spider leaderboard, as of August 2021, displays in the 1st rank position LGESQL~\cite{Cao2021} with Dev 0.751 and Test 0.720  for \emph{Exact set match without values}.
Thus RAT-SQL+GAP is arguably at the state-of-art in NL2SQL. 

RAT-SQL+GAP is based on the RAT-SQL package  (Relation-Aware Transformer SQL)
\cite{Wang2019}. RAT-SQL was proposed in 2019 as a text to SQL parser based on the
BERT language model \cite{Devlin2019}.
Package RAT-SQL version 3 with BERT is currently the 14th entry in the Spider leaderboard rank. 
RAT-SQL+GAP adds Generation-Augmented Pre-training (GAP) to RAT-SQL. GAP produces synthetic data to increase the dataset size  to improve pre-training; the whole generative models are trained by fine-tuning a BART \cite{Lewis2019} large model.

Despite the substantial number of techniques, systems and benchmarks for NL2SQL,
most of them focus on the English language. Very few
results can be found for input questions in the Portuguese language, for example.
The study by Silva et al. \cite{daSilva2021} presents an architecture for NL2SQL 
in which natural language questions in Portuguese are translated to the English 
language on arrival, and are then then shipped to NL2SQL existing packages.

The goal of this paper is simple to state: we present a translator for
queries in Portuguese natural language into SQL. 
We intend to study the effect of replacing the questions in the Spider 
dataset with translated versions, and also to investigate how to adapt the
RAT-SQL+GAP system to the needs of a different language. 
Using a new version of Spider with RAT-SQL+GAP to train models, 
we   produce inferences and   compare results so as to 
understand the difficulties and limitations of various ideas.

What we  found is that, by focusing on Portuguese,
we actually produced methods and results that apply to any multilingual
NL2SQL task. An important insight (and possibly the main
contribution of this paper) is dealing with a non-English
language, such as Portuguese, {\em we greatly benefit from taking a multilingual approach
that puts together English and the other language} --- in our case, 
English and Portuguese. We later stress this idea when we discuss
our experiments. We thus refer to our
``multilingual-ready'' version of RAT-SQL+GAP as mRAT-SQL+GAP; 
all code and relevant data related to this system are
freely available\footnote{mRAT-SQL+GAP Github: \url{https://github.com/C4AI/gap-text2sql}}.


\section{Preliminary Tasks}
The adaptation to language other than English demands at least the translation of the dataset and changing the code to read and write files UTF-8 encoding.

\subsection{Translating the Spider Dataset}

The translation to the Portuguese language in the NL2SQL task evolves the natural language part, the questions. The SQL queries must remain the same to make sense.
To translate the questions, it is important to extract them from specific .json files. The Spider dataset has three files that contain input questions and their corresponding SQL queries:
dev.json, train\_others.json, and train\_spider.json.
We extracted the questions and translated them using the Google Cloud Translation API.\footnote{Cloud Translation API: \url{https://googleapis.dev/python/translation/latest/index.html}.}.
Table~\ref{tab2} presents the number of questions and number of characters per file (just for the questions). 
The code that reads the original files and that generates translated versions relies on 
the UTF-8 encoding so as to accept Portuguese characters; several files were generated in the process (.txt just for the translated questions, .csv for the SQL queries and original/translated questions, .json for the translated questions).

\begin{table}[t]
\centering
\caption{Number of questions and characters per file.}\label{tab2}
\begin{tabular}{|l|l|l|}
\hline
File &  Number of questions & Number of characters\\
\hline
dev.json &  1,034 & 70,362\\
train\_others.json & 1,659 & 80,571\\
train\_spider.json & 7,000 & 496.054\\

\hline
\end{tabular}
\end{table}

We then conducted a  revision process by going through the text file and looking for
questions in the csv file (together with the corresponding SQL queries). 
After the revision, a new .json file was generated with these translated and revised
questions. Table~\ref{tab3} shows four examples of translated questions. 

\begin{table}
\centering
\caption{Translation examples.}\label{tab3}
\begin{tabular}{| p{6cm} | p{6cm} |}
\hline
English Question & Portuguese Question\\
\hline
How many singers do we have? & Quantos cantores nós temos?\\
Find the number of pets for each student who has any pet and student id. & Encontre o número de animais de estimação para cada aluno que possui algum animal de estimação e a identificação do aluno.\\
How many United Airlines flights go to City 'Aberdeen'? & Quantos voos da United Airlines vão para a cidade de 'Aberdeen'?\\
What is the name of the shop that is hiring the largest number of employees? & Qual é o nome da loja que está contratando o maior número de funcionários?\\
\hline
\end{tabular}
\end{table}

\subsubsection{Adapting the RAT-SQL+GAP System}

We had to change the RAT-SQL+ GAP code to allow multilingual processing. For instance, the original Python code is not prepared to handle UTF-8
files; thus, we had to modify the occurrences of ``open'' and ``json.dump'' commands, together with a few other changes. 
We ran a RAT-SQL+GAP test and checked whether all
the characters employed  in Portuguese were preserved.
We also noticed lemmatization errors in preprocessed files. As the original code for RAT-SQL+GAP
relies on the Stanford CoreNLP lemmatization tool that currently does not support
Portuguese, it was replaced by Simplemma.\footnote{Simplemma: a simple multilingual lemmatizer for Python at \url{ https://github.com/adbar/simplemma}.} 
The latter package supports multilingual texts, and particularly
supports Portuguese and English.


\subsubsection{Training}

The original language model at the heart of RAT-SQL+GAP is BART-large\footnote{Facebook BART-large: \url{https://huggingface.co/facebook/bart-large}.}~\cite{Lewis2019}, a
language model pretrained for the English language. 
We had to change that model to another one that was pretrained for the Portuguese language. 
A sensible option was to work with a multilingual Sequence-to-Sequence BART model; the choice was mBART-50 \footnote{Facebook mBART-50 many for different multilingual machine translations: \url{https://huggingface.co/facebook/mbart-large-50-many-to-many-mmt}}\cite{Tang2020} because it covers Portuguese and English languages (amongst many others).
Another language model we investigated was the BERTimbau-base\footnote{BERTimbau-base: \url{https://huggingface.co/neuralmind/bert-base-portuguese-cased}}\cite{Souza2020},
as RAT-SQL works with BERT; the move to BERTtimbau, a Portuguese-based version
of BERT, seemed promising.

\subsubsection{Dataset}

A total of 8,659 questions were used for training (7,000 questions in train\_spider.json and 1,659 questions in train\_others.json). The 1,034 questions in dev.json were used for testing. We later refer to three scenarios:
\begin{itemize}
\item English train and test: questions are just in English for training and testing.
\item Portuguese train and test\footnotemark: questions are just in Portuguese for training and testing.
\item English and Portuguese (double-size) train and test\footnotemark[\value{footnote}]: questions are in English and Portuguese for training and testing (in this case, we thus have twice as much data as in each of the two previous individual scenarios).
\footnotetext{Spider dataset translated to Portuguese and double-size (English and Portuguese together): \url{https://github.com/C4AI/gap-text2sql}}
\end{itemize}

\subsubsection{Evaluation Metrics}

The main evaluation metric with respect to Spider is Exact Set Match (ESM). 
We here present results for Exact Set Match (ESM) without values, as  
most results in the Spider leaderboard currently adopt this metric.
Some Spider metrics are also used to classify the SQL queries into 4 levels: easy, medium, hard and extra hard. Table~\ref{tab4} shows an example for each level; these queries correspond to the four questions in Table~\ref{tab3} in the same order.

To evaluate the results, we used the Spider test suite evaluation~\cite{Zhong2020}. 
An aside: the suite must receive a text file with the SQL query generated and another one
with the gold-standard SQL query. These files obviously do not change when we move
from English to any other input language. 
 
As a digression, note that it is possible to plug values during evaluation. A query with a value looks like this:
\begin{quote}
    SELECT Count(*) FROM airlines JOIN airports WHERE airports.City = \textbf{``Abilene"}
\end{quote}
A query without values has the word \textbf{``terminal"} instead of the value:
\begin{quote}
    SELECT Count(*) FROM airlines JOIN airports WHERE airports.City = \textbf{``terminal"}
\end{quote} 

\begin{table}[t]
\centering
\caption{SQL query levels: easy/medium/hard/extra.}\label{tab4}
\begin{tabular}{| l | p{11cm} |}
\hline
Level & SQL Query\\
\hline
Easy & SELECT count(*) FROM singer\\
Medium & SELECT count(*), T1.stuid FROM student AS T1 JOIN has\_pet AS T2 ON T1.stuid=T2.stuid GROUP BY T1.stuid\\
Hard & SELECT count(*) FROM FLIGHTS AS T1 JOIN AIRPORTS AS T2 ON T1.DestAirport = T2.AirportCode JOIN AIRLINES AS T3 ON T3.uid  =  T1.Airline WHERE T2.City = "Aberdeen" AND T3.Airline = "United Airlines"\\
Extra & SELECT t2.name FROM hiring AS t1 JOIN shop AS t2 ON t1.shop\_id = t2.shop\_id GROUP BY t1.shop\_id ORDER BY count(*) DESC LIMIT 1\\
\hline
\end{tabular}
\end{table}

\section{Experiments}

\begin{figure}[t]
\centering\includegraphics[width=0.90\textwidth]{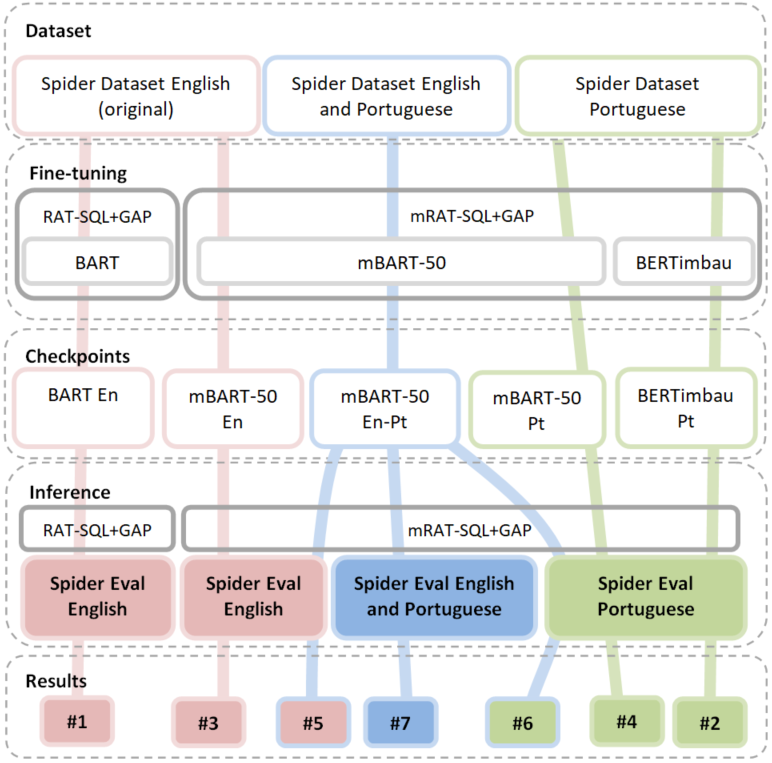}
\caption{Architecture of the training, inference, evaluation. Results related to Table~\ref{tab5}: each line in that table appears here as a square at the bottom 
of the figure.} \label{fig2}
\end{figure}

Experiments were run in a machine with 
AMD Ryzen 9 3950X 16-Core Processor, 64GB RAM, 2 GPUs NVidia GeForce RTX 3090 24GB running Ubuntu 20.04.2 LTS.
Fig.~\ref{fig2} shows the architecture of the training, inference and evaluation processes
described in this section. 

Results can be found in Table~\ref{tab5}. 
This table shows the results of Exact Set Match without Values for RAT-SQL+GAP 
trained locally for all models. 
We have 3 datasets, 5 trained model checkpoints and 7 distinct relevant results.

The first line corresponds to the original model BART and original questions in English. Note that the result in  line \#1  achieved the same performance reported by \cite{Shi2020} for  \emph{Exact Set Match without values} in Spider: 0.718 (All) for Dev. This indicates that our testing apparatus can produce state-of-the-art results.
Moreover, line \#1 shows a well-tuned model in English that can be attained. 

Our first experiment was to change the questions to Portuguese and the model to BERTimbau-base which is pretrained in Portuguese. In Table~\ref{tab5} the result in line \#2  for BERTimbau 0,417(All) is quite low when compared to the result in line \#1. 
This happens for many reasons. The model is BERT and the best result uses BART. Another important difference is that the Portuguese content used in fine-tuning has a mixture of Portuguese and English words because the SQL query inevitably consists of English keywords, see examples of SQL queries in Table~\ref{tab4}. In fact, some questions demand untranslated words to make sense, for example "Boston Red Stockings" in the translated question: \emph{Qual é o salário médio dos jogadores da equipe chamada "Boston Red Stockings"?}. This suggested that a multilingual approach might be more successful.
The mBART-50 language model was then tested within the whole architecture. 

mBART-50 was in fact fine-tuned in three different ways: with questions only in English, only in Portuguese, and with questions both in English and in Portuguese (that is, a dataset with questions and their translations). Inferences were run with English test questions and Portuguese test questions, while the model was fine-tuned with the corresponding training questions language. For the mBART-50 model fine-tuned with the train dataset with two languages, three inferences were made: only English, only Portuguese, and the combined English Portuguese test dataset.

\begin{table}[t]
\centering
\caption{Results.}\label{tab5}
\begin{tabular}{| c | c | c | c | p{1.3cm} | p{1.3cm} | p{1.3cm} | p{1.3cm} | p{1.3cm} |}
\hline
\multicolumn{9}{|c|}{\bfseries Exact Set Match without Values} \\
\hline
\# & Model & Train & Infer & Easy & Medium & Hard & Extra & \bfseries All\\
&  &  &  &  248 & 446 & 174 & 166 & \bfseries 1034\\
\hline
1 & BART & En & En & 0.899 & 0.744 & 0.667 & 0.428 & \bfseries 0.718\\
2 & BERTimbau & Pt & Pt  & 0.560 & 0.422 & 0.333 & 0.277 & \bfseries 0.417\\
3 & mBART-50 & En & En  & 0.851 & 0.679 & 0.546 & 0.386 & \bfseries 0.651\\
4 & mBART-50 & Pt & Pt  & 0.762 & 0.599 & 0.529 & 0.361 & \bfseries 0.588\\
5 & mBART-50 & En/Pt & En  & 0.863 & 0.682 & 0.569 & 0.422 & \bfseries 0.664\\
6 & mBART-50 & En/Pt  & Pt  & 0.827 & 0.596 & 0.511 & 0.331 & \bfseries 0.595\\
\hline
\hline
& Model & Train & Infer & Easy & Medium & Hard & Extra & \bfseries All\\
&  &  &  &  496 & 892 & 348 & 332 & \bfseries 2068\\
\hline
7 & mBART-50 & En/Pt  & En/Pt  & 0.847 & 0.639 & 0.537 & 0.380 & \bfseries 0.630\\
\hline
\end{tabular}
\end{table}

mBART-50 fine-tuned with questions in English in line \#3 achieved 0.651 (All) when tested with questions in English. The same model mBART-50 fine-tuned with questions in Portuguese in line \#4 achieved 0.588(All) when tested with questions in Portuguese.

A multilingual model such as mBART-50 can be trained with the two languages at the same time. This is certainly appropriate for data augmentation and to produce a fine-tuning process that can better generalize. The results in Table~\ref{tab5} lines \#5, \#6 and \#7 were obtained with mBART-50 fine-tuned with the double-size training dataset (English and Portuguese); the three inferences were made using the same model checkpoint. The test datasets were in English for line \#5, in Portuguese for line \#6, and the double-size test dataset in English and Portuguese for line \#7.
The results demonstrate improvements, if we compare inferences with English test dataset lines \#3 and \#5. Results went up from 0.651 (All) to 0.664 (All), better for all levels of questions. If we compare inferences with the Portuguese test dataset in lines \#4 and \#6, the results went from 0.588 (All) to 0.595 (All). However, they are better just for an easy level of questions; this was enough to influence the overall results for line \#6.

The inference made with the double-size test dataset in English and Portuguese, in line \#7, cannot be compared with the other inferences because they used just one language. Nevertheless, the model mBART-50 trained with English and Portuguese (double-size training dataset) produced good results with this rather uncommon testing dataset.

All the detailed results presented in this paper are openly available\footnote{mRAT-SQL+GAP Github: \url{https://github.com/C4AI/gap-text2sql}}.

\section{Analysis and Discussion}

These experiments indicate that multilingual pretrained transformers can be extremely useful when dealing with languages other than English. There is always the need to integrate English processing with the
additional languages of interest, in our case, with Portuguese.

\begin{figure}[t]
\includegraphics[width=\textwidth]{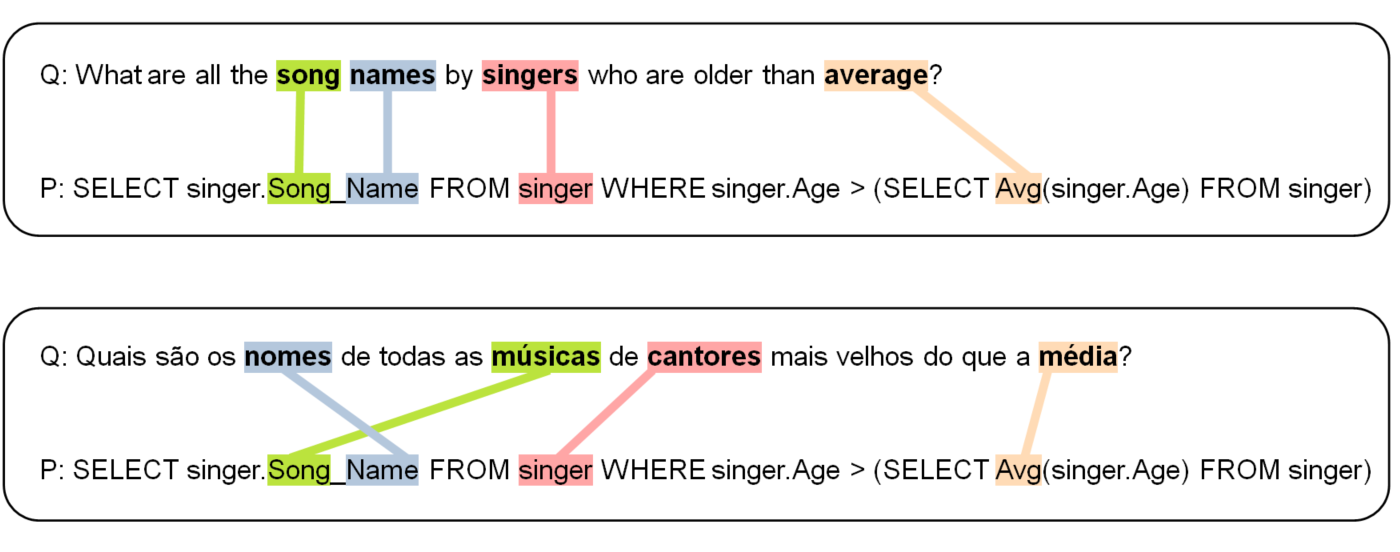}
\caption{Examples of keywords in the prediction of the SQL query in English and Portuguese languages. Top pair: question in English and corresponding SQL query predicted from it. Bottom pair: question in Portuguese and corresponding SQL query predicted from it.} \label{fig3}
\end{figure}

Overall, questions in the English language have a closer similarity
with SQL queries, thus simplifying inferences.
Conversely, questions in Portuguese require further work.
Fig.~\ref{fig3} show a real example of correct predictions in English (Table~\ref{tab5} line \#5) and in Portuguese (Table~\ref{tab5} line \#6).
In Fig.~\ref{fig3},  words such as \textbf{song, names, singers, average} in the English question are keywords needed to  resolve the query, 
and they are very close to the target word in the query. In Portuguese, the same keywords are \textbf{músicas, nomes, cantores, média} that must respectively match
song, name, singer, and avg. This introduced
an additional level of difficulty that  explains the slightly worse results 
for inference with  questions in Portuguese: 0.595 (Table~\ref{tab5} line \#6).
This is to be compared with  questions in English: 0.664 (Table~\ref{tab5} line \#5). In any case, it is surprising that the same model checkpoint resolved the translation with such different questions.

\begin{figure}[t]
\includegraphics[width=\textwidth]{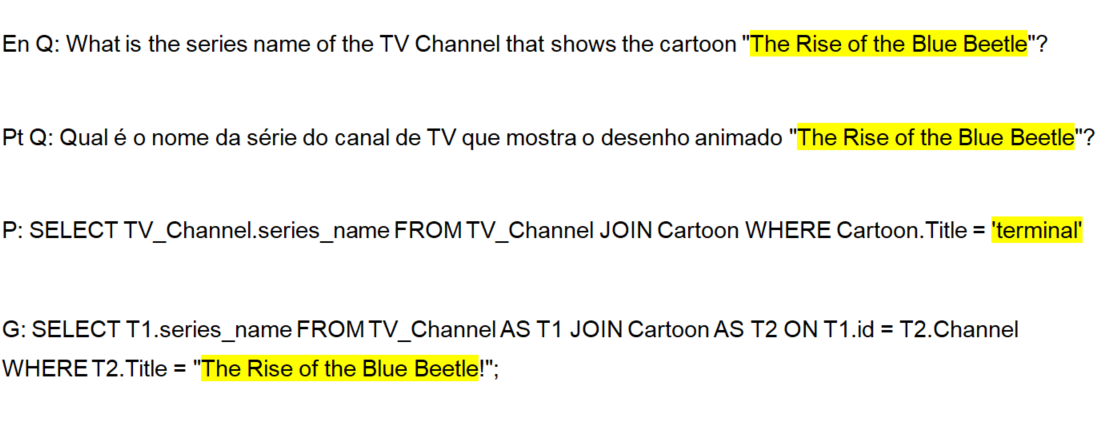}
\caption{Example of words that represent the value in the prediction of the SQL query. En Q: English question, Pt Q: Question translated to Portuguese, P: SQL query predicted, and G: Gold SQL query.} \label{fig4}
\end{figure}

Some translations actually keep a mix of languages, because some words represent the value and cannot be translated. Fig.~\ref{fig4} shows an example.
There the show name "The Rise of the Blue Beetle" should not be translated to maintain the overall meaning of the question; these words in English must be part of the Portuguese question. This successful SQL query inference was produced with mBART-50 fine-tuned with the English and Portuguese training dataset (Table~\ref{tab5} line \#6) and mBART-50 fine-tuned  in the Portuguese-only training dataset (Table~\ref{tab5} line \#4). The show name was then replaced with 
``terminal'' during the RAT-SQL+GAP prediction process, as it is processed through the Spider Exact Set Match \textbf{without Values} evaluation. In any case, the show name is part of the input and will introduce difficulties in the prediction.

In addition, for real-world databases, it is a practice, at least in Brazil, to name tables and columns with English words even for  databases with content in Portuguese. This practical matter is another argument in favor of a multilingual approach.

\begin{figure}[t]
\includegraphics[width=\textwidth]{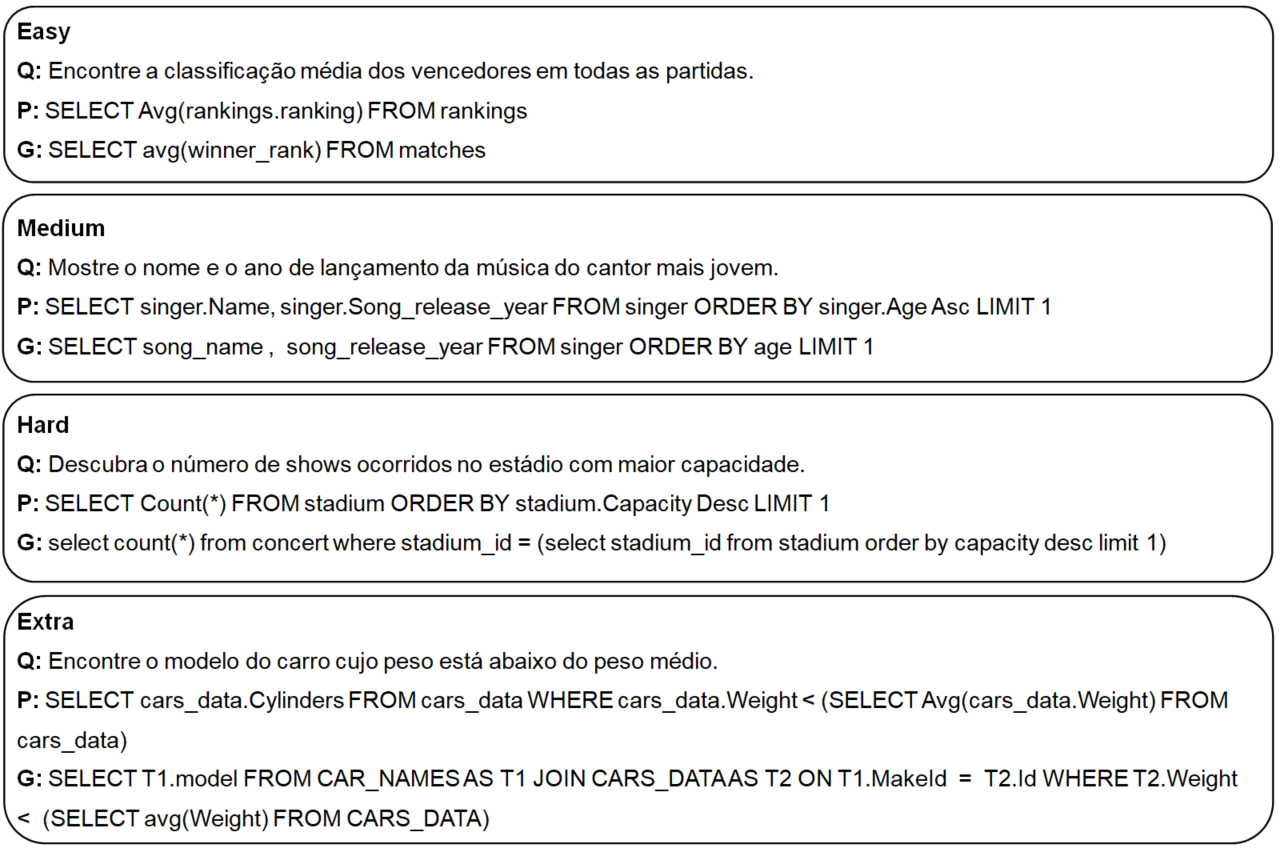}
\caption{Sample of failed queries evaluated by Spider Exact Set Match for inferences from mBART-50 (Table~\ref{tab5} line \#6) . Q: question, P: SQL query predicted, and G: Gold SQL query.} \label{fig5}
\end{figure}

Fig.~\ref{fig5} shows a sample of failed translations evaluated by the Spider Exact Set Match for inferences using mBART-50 (Table~\ref{tab5} line \#6).
It is actually difficult to find the errors without knowing the database schema related to every query. The objective of this figure is to show that even when the query is incorrect, it is not composed of random or nonsensical words. Our manual analysis indicates that this is true for queries failing with all other models.

\section{Conclusion and Future Work}

In sum, we have explored the possible ways to create a 
translator that takes questions in the Portuguese language
and outputs correct SQL queries corresponding to the questions.
By adapting a state-of-the-art NL2SQL to the Portuguese language,
our main conclusion is that a multilingual approach is needed:
it is not enough to do everything in Portuguese; rather, we must
simultaneously work with English and Portuguese.

In Table~\ref{tab5}, our best result is in line \#5 0.664 (All) whereby 
we test with questions in English (original test set from Spider) 
using mBART-50 model fine-tuned with a double-size training dataset 
(English and Portuguese). This yields 92\% of the English-only performance
of 0.718 (All) in  line \#1.
Testing with questions in Portuguese (using our translation) with the same  BART-50 model fine-tuned with a double-size training dataset (English and Portuguese), we achieve instead 0.595 (All) line \#6. Now this is 83\% of the English-only performance of 0.718 (All). These results should work as a baseline for future
NL2SQL research in Portuguese. 

Our multilingual RAT-SQL+GAP, or mRAT-SQL+GAP for short, the translated datasets, the trained checkpoint, and the results; are open-source available\footnote{mRAT-SQL+GAP Github: \url{https://github.com/C4AI/gap-text2sql}}.

Future work should try other multilingual transformers 
(and possibly other seq-to-seq models), always seeking ways
to use English and Portuguese together. 
Another possible future work is to fine-tune BERTimbau-large\footnote{BERTimbau-large: \url{https://huggingface.co/neuralmind/bert-large-portuguese-cased}}~\cite{Souza2020} so as to better understand the effect of the size of the language model.
Lastly, a translation of the Spider dataset to other languages so as
to work with several languages at the same time should produce valuable
insights.

\section{Acknowledgments}
This work was carried out at the Center for Artificial Intelligence (C4AI-USP), supported by the São Paulo Research Foundation (FAPESP grant \#2019/07665-4) and by the IBM Corporation. The second author is partially supported by the Conselho Nacional de Desenvolvimento Científico e Tecnológico (CNPq), grant 312180/2018-7.

%
%
%
%

\end{document}